\documentclass[10pt,letterpaper]{article}

\usepackage{cogsci}

\cogscifinalcopy 

\usepackage[utf8]{inputenc} 
\usepackage[T1]{fontenc}    
\usepackage{booktabs}
\usepackage{tabularx}
\usepackage{hyperref}
\usepackage{pslatex}
\usepackage{apacite}
\usepackage{float}
\usepackage{xcolor}         
\usepackage{graphicx}
\usepackage{adjustbox}
\usepackage{subcaption}
\usepackage{wrapfig}
\usepackage{layouts}
\usepackage[font=small]{caption}
\usepackage{color, colortbl}
\usepackage[group-separator={,},group-minimum-digits={3}]{siunitx}

\newcommand{\ignore}[1]{}

\newcommand{\ie}{\emph{i.e.,}}
\newcommand{\eg}{\emph{e.g.,}}

\newcommand{{\stcs}}{\emph{Constant Size}}
\newcommand{{\stca}}{\emph{Constant Area}}
\newcommand{{\stcac}}{\emph{Constant Area (contour)}}
\newcommand{{\stcc}}{\emph{Constant Circumference}}
\newcommand{{\stccc}}{\emph{Constant Circumference (contour)}}
\newcommand{{\stvs}}{\emph{Vary Size}}

\newcommand{\resnet}{\textsc{ResNet}}
\newcommand{\swin}{\textsc{Swin}}
\newcommand{\vit}{\textsc{ViT}}
\newcommand{\mlp}{\textsc{MLP}}

 \title{Evaluating Visual Number Discrimination in Deep Neural Networks}
 
\author{{\large \bf Ivana Kaji\'c (kivana@deepmind.com)} \\
  DeepMind, Montr\'eal, QC, Canada \\
  \AND {\large \bf Aida Nematzadeh (nematzadeh@deepmind.com)} \\
  DeepMind, London, United Kingdom
  }

\begin{document}

\maketitle

\begin{abstract}

The ability to discriminate between large and small quantities is a core aspect of basic numerical competence in both humans and animals. In this work, we examine the extent to which the state-of-the-art neural networks designed for vision exhibit this basic ability. Motivated by studies in animal and infant numerical cognition, we use the numerical bisection procedure to test number discrimination in different families of neural architectures.
Our results suggest that vision-specific inductive biases are helpful in numerosity discrimination, as models with such biases have lowest test errors on the task, and often have psychometric curves that  qualitatively resemble those of humans and animals performing the task.
However, even the strongest models, as measured on standard metrics of performance, fail to discriminate quantities in transfer experiments with differing training and testing conditions, indicating that such inductive biases might not be sufficient.

\ignore{
The ability to discriminate large and small quantities is a core aspect of basic numerical competence in both humans and animals. In this work, we examine the extent to which the state-of-the-art neural networks designed for vision exhibit this basic ability. Motivated by studies in animal and infant numerical cognition, we use the numerical bisection procedure to test number discrimination in different families of neural architectures.
Our results suggest that vision-specific inductive biases are helpful in numerosity discrimination, as models with such biases have lowest test errors on the task, and often have psychometric curves that  qualitatively resemble those of humans and animals performing the same task.
However, even the strongest models, as measured on standard metrics of performance, fail to discriminate quantities in transfer experiments with differing training and testing conditions, indicating that such inductive biases might not be sufficient.
}

\end{abstract}

\section{Basic Numerical Competence}
The ability to represent abstract numbers and compare numerical quantities is a basic numerical competence observed in both animals and humans~\cite{dehaene1998abstract}.
%
It helps animals in foraging, navigation, hunting, and reproduction~\cite{nieder2020adaptive}, and is also correlated with the later mathematical ability in prelinguistic infants~\cite{gilmore2007symbolic,halberda2008individual}.
While such a skill is shared across species and is independent of explicit feedback or formal education~\cite{dehaene2011number,gallistel1992preverbal},
the degree to which more advanced numerical skills, such as counting and symbolic representation of number, are present across species remains a debated topic~\cite{o2021cultural,revkin2008does,anobile2016number,gallistel1992preverbal}.

To investigate number representation and processing, different neural networks have been used as cognitive models of various numerical skills such as magnitude comparison~\cite{verguts2004representation,dehaene1993development,Zorzi1999-ld}, subitizing~\cite{peterson2000computational} and counting~\cite{rodriguez1999recurrent,fang2018can}.
Neural networks are able to encode exact magnitudes~\cite{CREATORE2021104815} and develop basic numerical abilities such as numerosity comparison~\cite{Testolin2020-tg}.

While such networks have been used successfully to explain different phenomena in numerical cognition, their architecture is often designed for a task targeting specific cognitive function.
In contrast to such specialized networks, in recent years we have witnessed a radical improvement in both the performance, and the quality of representations learned by deep neural networks that are trained end-to-end across vision~\cite{simonyan2014very,he2016deep}, language~\cite{vaswani2017attention,devlin2018bert,brown2020language}, and multimodal~\cite{lu2019vilbert,radford2021learning,alayrac2022flamingo} domains.

Here, we investigate whether state-of-the-art models designed for visual processing, also referred to as vision encoders, can exhibit basic numerical competence as observed in humans and animals. 
Specifically, we evaluate \emph{number discrimination} in vision encoders, defined as the ability to make broad relative numerical judgements such as many versus few, which is imprecise and not as advanced as counting, but within the normal ability of many animals~\cite{Davis1982-lv}.
We draw inspiration from studies in animal and child cognition and use a simple discrimination paradigm known as the \emph{bisection task} to examine if recent vision encoders can learn to discriminate stimuli on the basis of number.
%

We consider three vision encoders with varying degrees of explicit inductive biases: \resnet~\cite{he2016deep}, \vit~\cite{dosovitskiy2020image}, and \swin~\cite{liu2021swin}, as well a simple, comparatively small,  multi-layer perception (\mlp{}) not designed for vision tasks as a baseline.
%
Across all conditions, \swin{} and \resnet{} with image-specific inductive biases
are the most successful models in number discrimination; moreover, \swin{} matches the empirical data from humans and animals in more conditions than \resnet{} suggesting that its additional hierarchical bias results in a better abstract number representation.
Even the strongest models, however, often fail in conditions that test for the transfer of numerical skill to a new condition; for example, when models are trained on a stimulus with solid shapes but tested on a stimulus where shapes are not filled. 
Although models fail in such transfer conditions, we find that they do learn structured number representations, forming clusters that are ordered based on the number identity. This suggests that, unlike humans and animals whose numerical skills generalize across different ecological contexts, vision encoders might require additional modeling innovations or a greater quantity and variety of data to use their learned knowledge in new situations.
\section{The Numerical Bisection Task}
\label{sec:stimuli}
The numerical bisection task is used to assess perception of numerical quantitites in both animals and humans.
First, a participant is trained to discriminate small and large sample numerosities by associating them with different responses (labels such as \emph{few} and \emph{many}). For example,~\citeA{Emmerton1997-gj} train pigeons to respond to images with 1 or 2 shapes by pecking to the left (corresponding to \emph{few}), and to the right for images with 6 or 7 shapes  (corresponding to \emph{many}). In~\citeA{Almeida2007-lz}, children learn to pick a green cup for 2 drumbeats, or a blue cup for 8 drumbeats in one experiment, and raise a red glove on their left hand after 2 drumbeats and a yellow glove on their right hand after 8 drumbeats in another experiment. The numerosities used for training (\ie{} 1, 2, or 8) are often referred to as \emph{anchor numerosities}.

Then, to probe number discrimination, participants are subsequently tested on intermediate numbers that are \emph{not} seen during training (\eg{} 3 in the previous experiment). A participant is more likely to select the response associated with the larger anchor value (\eg{} \emph{many}), resulting in an s-shaped psychometric curve.
Such s-shaped psychometric curves have been used to characterize basic numerical competence in rats~\cite{Meck1983-gq}, pigeons~\cite{Honig1989-dj,Emmerton1997-gj}, rhesus macaques~\cite{jordan2006weber}, as well as children and adults~\cite{Droit-Volet2003-ts,Almeida2007-lz,jordan2006weber}.
Qualitatively, psychometric curves documented in the literature have the following characteristics: (1) the initial segment with smaller numerosities is mostly labeled with \emph{few}, (2) intermediate segment with a gradually increasing slope reflecting an increase in \emph{many} responses, (3) final segment with the largest numerosities mostly labeled with \emph{many}. Although these properties characterize the majority of psychometric responses documented in the literature, between- and within-subject variability has been observed depending on the task and numerosity ranges  ~\cite{Almeida2007-lz}.


\subsection{Experimental Stimuli}
\begin{figure}[t]
\centering
\includegraphics[width=.45\textwidth]{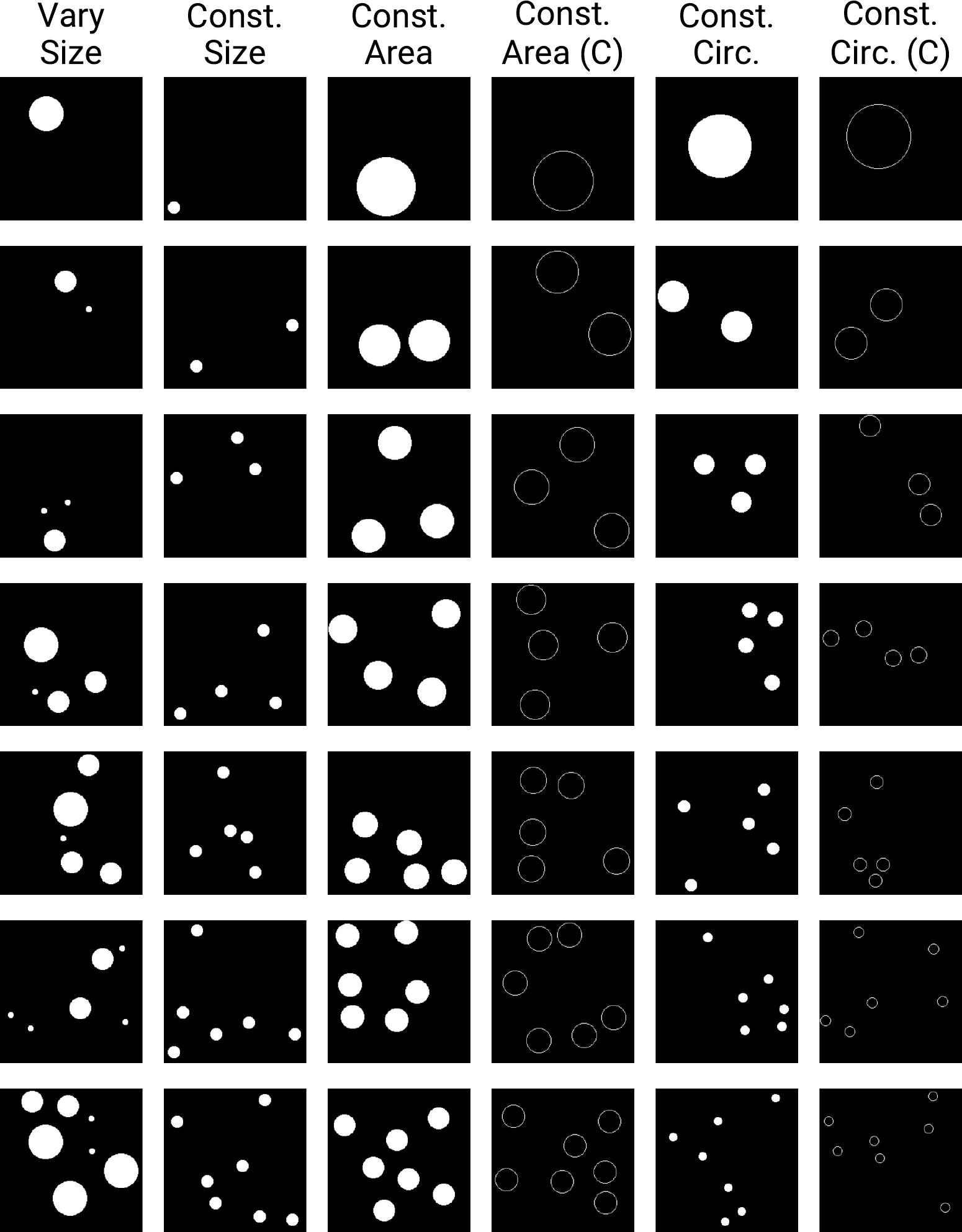}
\caption{
Sample visual stimuli used in the numerical bisection task. Rows are  different numerosities and columns different stimuli types. "(C)" denotes "contours", as opposed to shapes with solid white background.
}
\label{fig:stimuli}
\end{figure}
%
We automatically generate images with black background and white circles varying the number of circles from 1 to 7. Similar to~\citeA{Emmerton1997-gj}, we use images with 1, 2, 6, or 7 circles as anchor numerosities for training.
When designing stimuli, previous work has identified and controlled for potential perceptual confounds such as the size of the constituent elements (\ie, circles in our case), total white area, or total perimeter~\cite{Honig1989-dj,Testolin2020-tg,Emmerton1997-gj};
%
processing these non-numerical features---which may be a confound in the observed numerical discrimination behavior---can develop independently of number processing, as has indeed been observed in children's developmental trajectory~\cite{odic2018children}.
%
To control for such potential confounds, we generate six different stimulus categories shown in Figure~\ref{fig:stimuli}:
\begin{enumerate}
     \item \stvs{}. This is our most general setting, where for each image, we draw circles with radii drawn randomly from a set of 3 values ($r=\{10, 35, 55\}$).
    \item \stcs{}. We control for the size of circles---all circles have the same radius ($r=20$); this enables us to examine whether models can discriminate numbers better when circles are identical compared to varied in size.
    \item \stca{}. In the previous condition (\stcs), the white area (covered by circles) increases as the number of circles increases. We control for this potential confound by fixing the total white area to be constant across stimuli. This results in smaller circles in images that depicts larger numbers. 
    \item \stcac{}. We also examine if solid shape background  has an impact on models' behavior; we consider a condition the same as \stca, but using contours instead of shapes with white background.
    \item \stcc{}. While the total area is controlled for in the \stcs{} condition, the total circumference of circles increases with numerosity. Here, we control for the total circumference by keeping it constant across stimuli.
    \item \stccc{}. It is the same as \stcc, but using contours instead of shapes with white background.
  
\end{enumerate} 

We generate the stimuli on-the-fly for both the training and testing and store them in-memory to be used during training and testing, with 100 images generated for each numerosity category of one stimulus type, resulting in overall \num{400} images for training, and \num{1100} images for testing for each stimulus type.
The images are of dimensionality expected by models, \ie{} $224\times224\times3$.


\section{Experimental Setup}
\label{sec:methods}
In this section, we examine a few recent families of deep neural networks designed for computer vision (henceforth, vision encoders); all these models have achieved impressive results on computer vision tasks (such as image classification), but differ with respect to inductive biases that their architecture encode.  We first describe these models briefly, and then discuss the details of our experimental setup.

\paragraph{Models.}
We consider three types of vision encoders: ResNet~\cite{he2016deep}, \vit~\cite{dosovitskiy2020image}, and Swin ~\cite{liu2021swin}.
%
%
The ResNet model includes a stack of convolutional neural network (CNN) blocks that process images using convolution kernels. These kernels introduce an explicit \emph{locality} bias---pixels (or features depending on the layer) that are close spatially are combined; as a result, a model with CNN blocks typically learns to encode low-level features (such as edges) in its first layers, and more high-level ones (such as parts) in its last layers.

Both \vit{} and \swin{} use Transformer blocks~\cite{vaswani2017attention} consisting of feed-forward layers and a \emph{self-attention} mechanism; self-attention introduces a weaker and less explicit \emph{locality} bias (compared to CNNs) as a model can learn to group neighboring image patches.\footnote{Self-attention is designed for sequential data such as language; thus, it is less suitable for modelling the two-dimensional spatial relations among image patches.}
\swin{} builds on \vit{} and introduces an explicit \emph{hierarchical} bias by modifying how self-attention is applied across different layers; more specifically, local image patches are merged at at various stages as the depth of the model increases, resulting in a hierarchical representation.
%

We use specific variants of \resnet{}, \vit{}, and \swin{} encoders: the ResNet-50 variant with 25.6M parameters~\cite{he2016deep}, \vit-B~\cite{dosovitskiy2020image} with 86M parameters, and ``tiny'' Swin, \swin-T~\cite{liu2021swin}, with 29M parameters. We picked the smallest \vit{} and \swin{} variants, and a \resnet{} model that has a similar number of parameters to \swin{}.

Finally, as a simple baseline, we consider a generic feed-forward multi-layer perceptron (MLP) that does not include any inductive biases such as convolutions or attention which are known to be helpful for processing of real-world images. 
We use an MLP consisting of 2 hidden layers with 256 units each, separated by ReLU non-linearities, and a final linear layer with 2 units. With 0.13M parameters and no ``bells and whistles'', this makes it a substantially smaller, yet less computationally inexpensive baseline model.

\paragraph{Training.} For each stimulus type (\eg{} \stca), we train \resnet{}, \vit{}, \swin{} and \mlp{} models on data generated for that stimulus, \ie{} images and their labels (\emph{few} and \emph{many}).
More specifically, we add a classification head to these models, to predict the label \emph{few} for images with 1 and 2 circles, and \emph{many} for images with 6 and 7 circles, where labels are encoded as one-hot vectors.
All models are trained with a cross-entropy loss and L2 regularization.
To get an estimate of variability in model responses for each stimulus category, we train 10 networks by choosing a different seed that randomly initializes network weights.

We perform a hyper-parameter search on the batch size, number of steps, learning rate, and optimizer type to find combinations where training loss has converged on the validation set, and where a network is achieving close to 100\% accuracy on the training set.  
Accuracy is defined as a percentage of correctly classified labels.\footnote{
We find that the batch size of 16, and \num{5000} steps worked well for all models, although losses in some models (\eg{} \swin{} and \resnet) converged much faster. We use the Adam optimizer~\cite{kingma2014adam} for \mlp{}, \vit{}, and \swin{} models, with learning rates of 1e-04, 5e-04, and 5e-05, respectively. We use the SGD optimizer with a learning rate of 1e-2 for \resnet. The models are trained using a NVidia Tesla V100 GPU.}


%
\paragraph{Testing.} 
We test the models on new images of anchor numerosities (\ie{} \emph{not} seen during training), as well as images of novel interpolated numerosities: 3, 4 and 5. 
We use 100 images for each numerosity category and each stimulus type.

\section{Experimental Results}
\label{sec:xps}
In Experiment 1, we investigate models' behavior when trained and tested on the same stimulus type. In Experiment 2, we investigate transfer of the number discrimination skill by testing models on images from a stimulus category that is not used in training (\ie{} train on \stcs, and test on \stca).

\subsection{Experiment 1: Number Discrimination}
In this experiment, we test number discrimination using images from the same stimulus category that is used during training (\eg{} train on \stcs, and test on \stcs). 
We evaluate models based on the accuracy of the stimulus test set, and the quantitative and qualitative characteristics of psychometric curves in relation to the empirical data.

\begin{table*}[tp]
\centering

\scalebox{0.95}{
\begin{tabular}{l|rrrr|rrrr|rrrr}
\toprule
{} & \multicolumn{4}{c}{Few (1, 2)} & \multicolumn{4}{c}{Many (6, 7)} & \multicolumn{4}{c}{Total Error (Few+Many)} \\
{}  & \resnet &  \vit & \swin &  \mlp  & \resnet &  \vit & \swin &  \mlp & \resnet &  \vit & \swin &  \mlp  \\
\midrule
Vary Size         &    3.0 & 11.8 &  0.2 & \textbf{17.4} &    3.6 & \textbf{11.9} &  0.6 &  7.8 &    3.3 & 11.8 &  0.4  &        \textbf{12.6} \\
Const. Size       &    0.0 & \textbf{35.5} &  0.0 & 32.1 &    0.0 &  0.3 &  0.1  &   \textbf{5.0}  &    0.0 & 17.9 &  0.0 &        \textbf{18.6}\\
Const. Area       &    1.1 & \textbf{33.0} &  0.0 &  7.5 &    0.8 &  \textbf{7.2} &  0.1 &  2.8   &    0.9 & \textbf{20.1} &  0.1 &         5.1 \\
Const. Area (C)  &    0.4 &  8.0 &  0.0  & \textbf{63.2} &    1.5 & \textbf{12.8} &  0.3 & 10.1 &    0.9 & 10.4 &  0.1  &        \textbf{36.6}\\
Const. Circ.      &    0.6 &  0.4 &  0.0 &  3.5  &    0.0 &  3.5 &  0.0 & \textbf{51.8} &    0.3 &  1.9 &  0.0  &        \textbf{27.6}\\
Const. Circ. (C) &    8.7 & \textbf{40.8} &  0.7 & 31.1 &    9.9 & 35.0 &12.8  & \textbf{44.6}  &    \underline{9.3} & \underline{\textbf{37.9}} &  \underline{6.8} &   \underline{\textbf{37.9}} \\
\bottomrule
\end{tabular}
}
\caption{Error rates (\%) in classifying anchor numerosities as either ``few'' or ``many'' on respective test sets. Highest error rates for each stimulus type and each anchor numerosity are highlighted. Highest total error rates across stimuli for each model are underlined.}\label{tab:results}

\end{table*}

\paragraph{Performance on seen numbers.} We first examine the performance of the four architectures when tested on novel images of the anchor numerosities (seen during training): 1, 2, 6, 7. An error occurs when an image with a small numerosity (\ie{} 1 or 2 circles) is classified as \emph{many}, or when an image with a large numerosity (\ie{} 6 or 7) is classified as \emph{few}.
Average error rates for each network and each stimulus type are shown in Table~\ref{tab:results}.
Overall, we observe that \resnet{} and \swin{}, with image-specific inductive biases, have smallest mean error rates of less than 1\% in 6/12 and 11/12 conditions, respectively.
\vit{} has mean error rates that are in some cases comparable to or even exceed errors of the \mlp{} baseline.
When averaged across all 4 numerosities, we find that highest error rates are consistently observed with the \stccc{} stimulus category (See Table~\ref{tab:results}, column ``Total error''); suggesting that this combination of visual features represented the most challenging dataset for number abstraction.
Meanwhile, no such consistent pattern exists for datasets resulting in smallest errors---the smallest error for \resnet{} is observed with \stcs, for \vit{} and \swin{} with \stcc, and for the \mlp{} with \stca.
This observation is not surprising given that these models encode different inductive biases.

\begin{figure}[t]
    \centering
    \includegraphics[width=.47\textwidth]{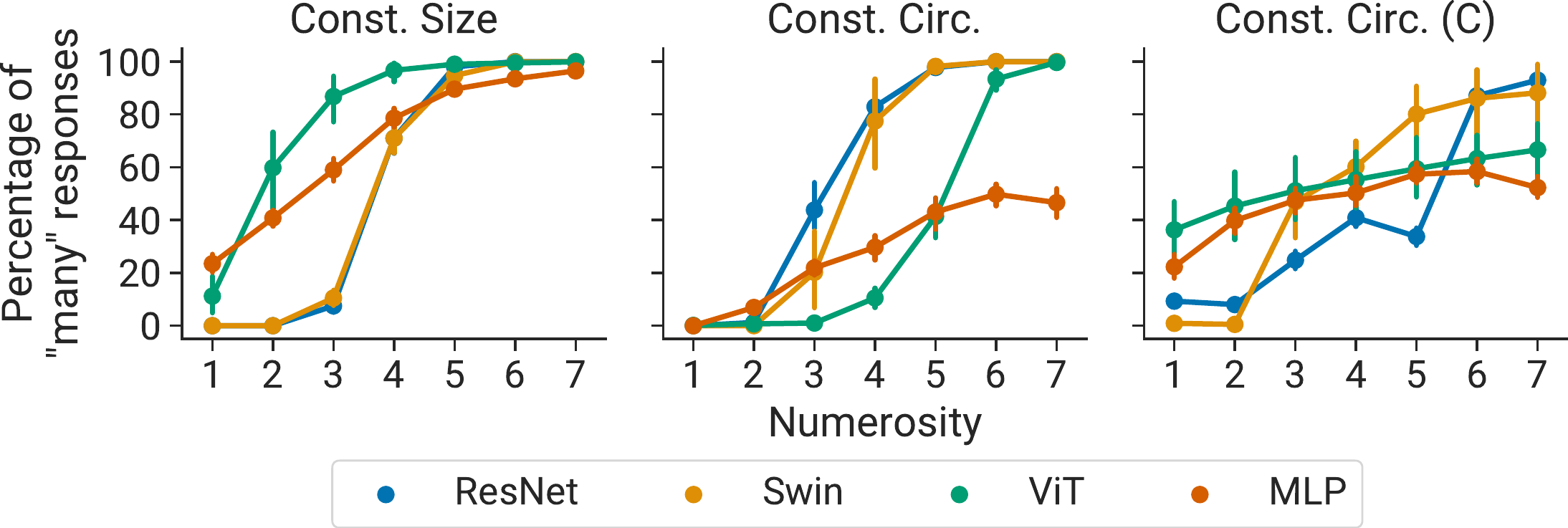}
    \caption{
    Psychometric functions for each model trained and tested on one stimulus type on the numerical bisection task. Vertical bars are 95\% bootstrapped CIs.
    }
    \label{fig:psych}
\end{figure}

\paragraph{Performance on new numbers.} Next, we examine how different models perform on numbers not seen at the training time (\ie{} 3, 4, and 5). We plot the psychometric curves for selected stimuli, showing percentages of \emph{many} responses across numerosities for models trained on that stimuli in Figure~\ref{fig:psych}.
We selected these stimulus categories as representative of the easiest (\stcs, \stcc) and hardest (\stccc) conditions based on the average error rates in Table~\ref{tab:results}.
Different from Table~\ref{tab:results}, each value on the y-axis represents a proportion of \emph{many} responses for a certain numerosity (x-axis).

Overall, some curves in Fig.~\ref{fig:psych} exhibit characteristics of typical psychometric functions as discussed in Sec.~\nameref{sec:methods}---specifically, for small numerosities 1, 2, and sometimes 3, we observe a slowly accelerating initial segments, followed by gradual increase with intermediate numbers, and a slowly decelerating final segment for larger numerosities (6, 7). Examples of such curve profiles are \swin{} and \resnet{} responses to \stcs, and \stcc{} stimulus categories.
However, there are also curves that have atypical flat shapes indicative of failure to learn this task, \ie{} those not found in the literature. Out of all 24 curves we analyzed, 4 in total exhibit such a shape with 3 shown in Fig.~\ref{fig:psych} (\ie{} \mlp{} with \stcc{} and \mlp{} with \stccc, \vit{} with \stccc). Such curves either do not start at, or do not end at expected values indicating lack of sensitivity to number categories, some of which is also evident based on large error values in Table~\ref{tab:results}.


\subsection{Experiment 2: Transfer to Novel Stimuli}
\begin{figure}[t]
\centering
\includegraphics[width=.47\textwidth]{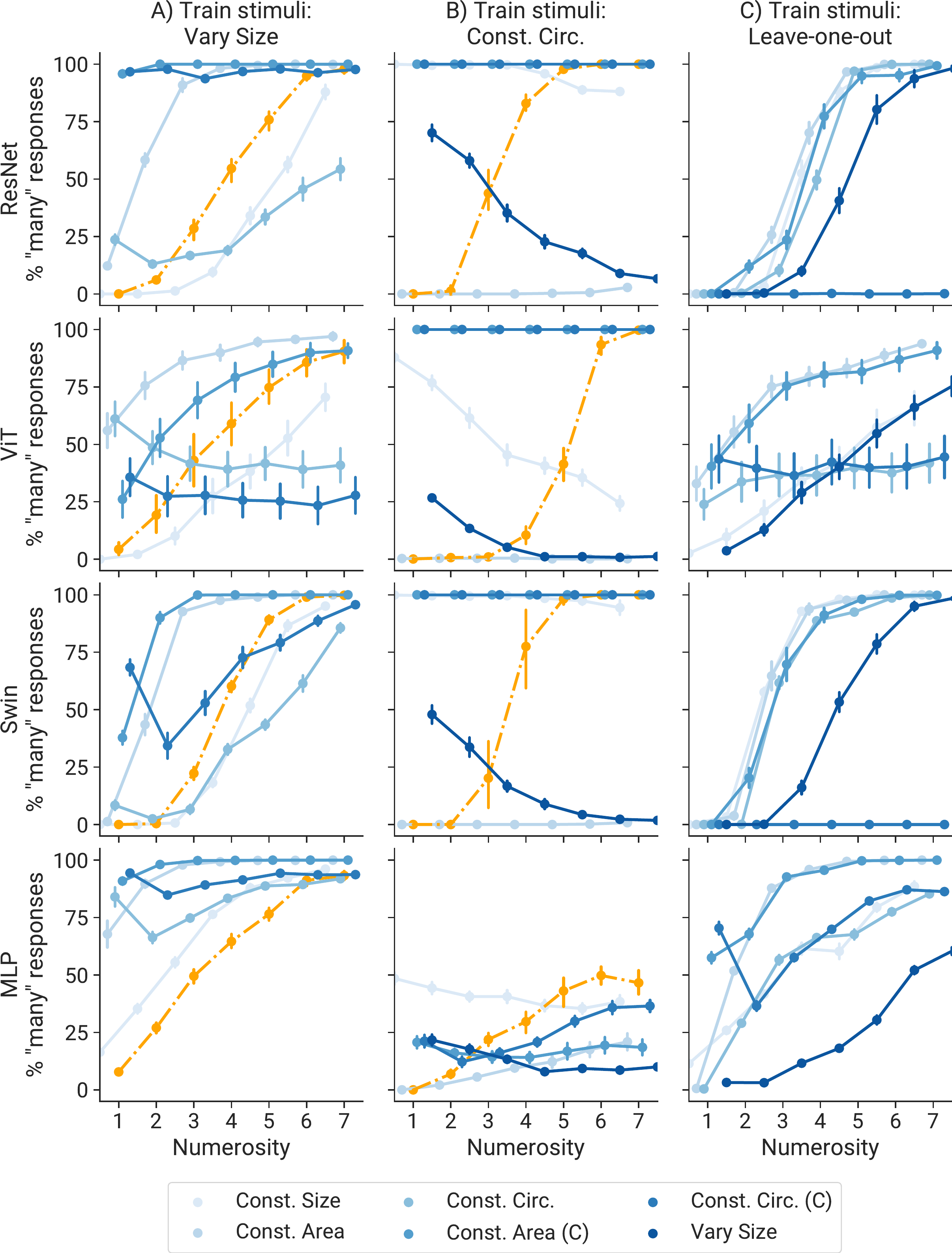}
\caption{
Psychometric curves in transfer experiments. Bars are standard errors of the mean. Models are represented across rows, and selected training datasets in columns. Shades of blue indicate the test stimulus, orange curves denote train and test on the same stimulus type.
}
\label{fig:transfer}
\end{figure}
The previous experiment demonstrates that \swin{} and \resnet{} architectures to a great degree appear to be able to  differentiate number of items in an image when trained and tested on the same stimulus type (\eg{} constant total area or circumference).
To understand whether our models have indeed developed a notion of a number category as opposed to learning a given stimuli, we draw a parallel with research in animal cognition and examine if the models ``\emph{base their behaviour on the numerosity of a set, independent of its other attributes}''~\cite{gallistel1992preverbal}.
In other words, if models learn an abstract representation of a number category, we would expect this representation to be agnostic to perceptual features of the stimulus. 
To test this, we examine models in a cross-stimulus transfer setting: we train a model on one set of stimuli, but test it on other types of stimuli (\ie{} train on \stcs, test on \stcc). The test stimuli are \emph{out of distibution (OOD)} with respect to the model's training distribution. Compared to the \emph{in distribution} setting where training and test are drawn from the same distribution, the OOD setting is known to be challenging for neural networks~\cite{geirhos2020shortcut}.\footnote{The degree to which a stimulus category is OOD depends on the target distribution, as some stimulus categories are correlated across different dimensions; \eg{}, total cumulative white area increases on average with numerosity for both \stcs{} and \stvs.}

Figures~\ref{fig:transfer}A) and ~\ref{fig:transfer}B) show a selected subset of psychometric curves evaluated using such a cross-stimulus protocol, for models trained on \stvs{} and \stcc. 
Orange curves denote cases where a model is trained and tested on the same stimulus category (\ie{} the protocol from Exp. 1), and are included for reference, while blue curves are obtained when models are tested on datasets different from the ones they are trained on.
We report results on \stvs{} and \stcc{} since they exhibit the most 
successful (\stvs) and least successful (\stcc) transfer cases, as defined by the expected qualititative characteristics of psychometric curves in Sec.~\nameref{sec:stimuli}.
Even for the best matching condition \stvs{} (Fig.~\ref{fig:transfer}A), we observe a number of \emph{transfer failures}, where a trained model shows poor transfer of number discrimination ability to a novel stimulus category, revealing a failure to abstract the number category.

An interesting case of transfer failures is evident in \emph{all-or-none} responses, where models unanimously assign either \emph{few} or \emph{many} response to all numerosities. This has been observed across all models, and is particularly prominent with models trained on \stcc{} (Fig.~\ref{fig:transfer}\,B). In some cases, most notably with \mlp{} and to a smaller extent with \vit, we also observe flatter curves with smaller slopes, resulting from frequent misclassification of \emph{many} responses as \emph{few} and vice versa. Finally, we also observe a new response pattern, an \emph{inverted} psychometric curve, where small numerosities are overwhelmingly assigned label \emph{many}, and the opposite for large numerosities. Fig.~\ref{fig:transfer}B) showcases that this pattern is consistent across models trained on \stcc. We conjecture that this is due to models latching onto total white area during training, which is inversely correlated with numerosity in \stcc.

Next, we consider an easier case of transfer, and examine if exposing models to more a diverse set of stimuli (as opposed to one type of stimulus) can help in learning a better representation of number categories; we train models on all but one stimulus type, and evaluate them on the hold-out stimulus type. 
Instead of 100 images per number category, a model is seeing 500 images per number category (\ie{} 100 images for a number for each of the 5 training stimulus types).
%
%
As shown in Fig.~\ref{fig:transfer}C), we see that increasing data variability results in more curves that resemble the expected s-shaped curve, especially for \resnet{} and \swin{}. However, even then, models failed to generalize to \stccc, confirming again the difficulty of this stimulus category.
Overall, we find that \swin{} and \resnet{} produce representations that better match observed empirical data even in the more challenging transfer setting. We also observe that the training models on a variety of stimulus types help in generalising to new stimulus.

\begin{figure}[t]
\centering
   \begin{subfigure}{0.49\linewidth} \centering
     \includegraphics[scale=0.35]{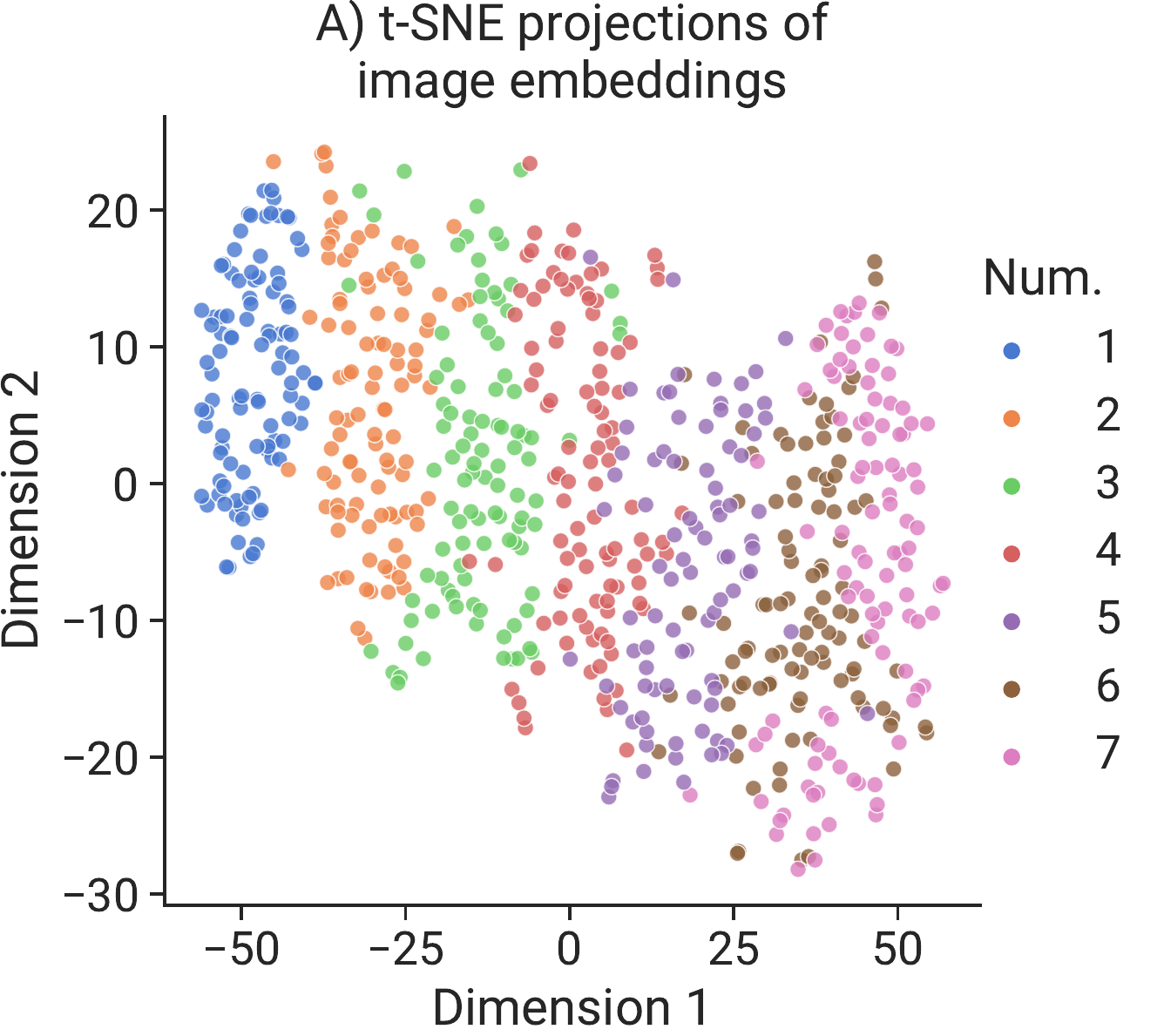}
   \end{subfigure}
   \begin{subfigure}{0.49\linewidth} \centering
     \includegraphics[scale=0.35]{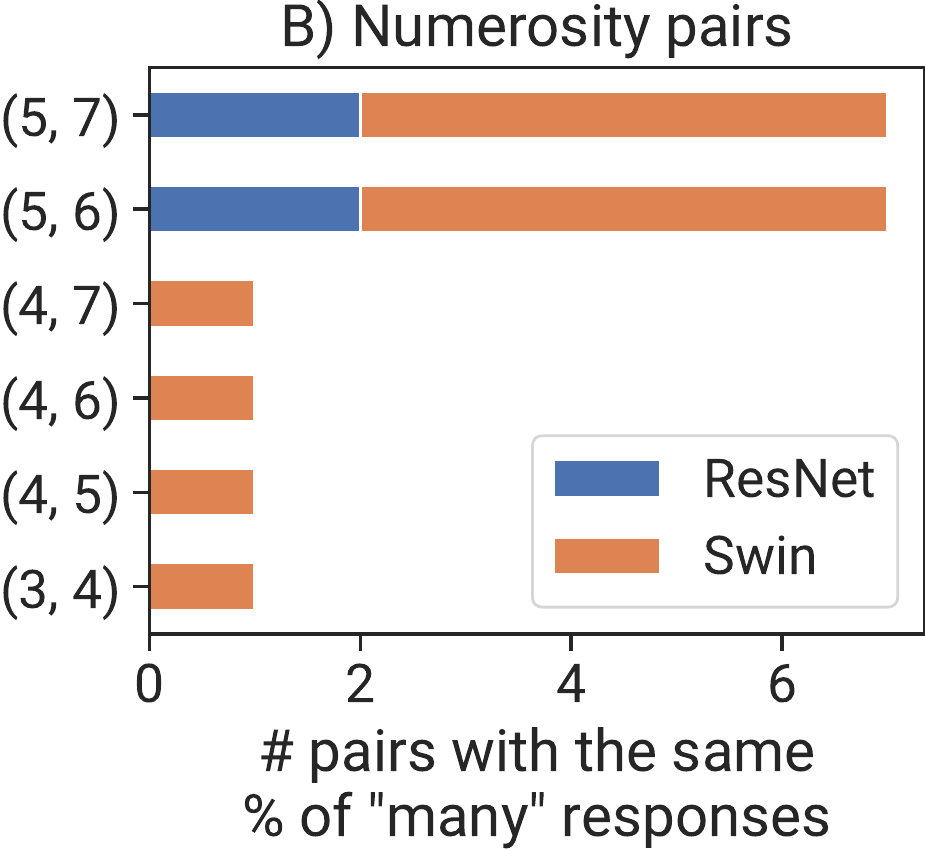}
   \end{subfigure}
\caption{A) t-SNE projections of \stcs{} image embeddings for \swin{} trained on \stca. B) Numerosities of image pairs for which there was no significant difference in the percentage of ``many'' responses. }\label{fig:twofigs}
\vspace{-0.1cm}
\end{figure}

Finally, we examine if learned number representations form meaningful clusters; to answer this question, we do a forward pass on images from a given stimulus category for two models with lowest error rates (\ie{} \resnet{} and \swin). For each image, we extract embeddings from the last dense layer of the model, prior to the 2-unit classification head. We use PCA followed by the t-SNE~\cite{van2008visualizing} dimensionality reduction method to project high-dimensional embedding vectors (\num{2048} for \resnet{} and 768 for \swin) into 2D space. In Fig.~\ref{fig:twofigs}A) we show one selected example of such a projection, where individual points have been color-coded based on the numerosity of the stimulus image.
First, we observe that embeddings cluster in groups based on number, with a greater cluster overlap for subsequent numbers. Second, we observe an ordering of clusters based on numerosities. This type of pattern is observed more often with embeddings from \swin{}, compared to \resnet{} embeddings which generally result in less discernible clusters (with the exception of clusters for numerosities 1 and 2). Interestingly, based on visual inspection of the data, we do not find that more distinct projections suggest better performance on the task. For example, while clusters in Fig.~\ref{fig:twofigs}A) seem to be discernible based on number, the model performs poorly when tested on \stcs, possibly because the classifier does not discriminate based on the dimensions that are discriminable in the embeddings.


\section{Discriminability of Small vs. Large Numbers}

Empirical data from humans and animals shows that in the numerical bisection task, it is consistently more difficult to distinguish larger numerosities from each other, compared to  smaller ones~\cite{Almeida2007-lz,Emmerton1997-gj}.
This observation is likely to be related to a more general finding in numerical cognition that small numbers are processed differently than larger numbers~\cite{dehaene2011number,revkin2008does}.

We examine whether similar observations can be made for our models’ responses, using a measure of discriminability for different pairs of numbers. 
As an example, for a given model and a pair of numbers (such as 5 and 6), we statistically test if the mean percentage of \emph{many} responses for images of one number (5) is the same as that of images of the other number (6). Intuitively, the two numbers are harder to discriminate if their mean percentage of \emph{many} responses are the same.
We consider models with smallest test error rates in Experiment 1 (\ie{} \resnet{} and \swin).
For a given model and each stimulus type, we consider all possible number pairs (\ie{} all points on average psychometric curves) and perform the Tukey’s HSD test for multiple pair-wise comparison of means (with family-wise error rate FWER=$.05$). This approach is based on the similar statistical tests used with pigeon responses in ~\citeA{Emmerton1997-gj}.

In Fig.~\ref{fig:twofigs}B), we show the breakdown of 18 cases (out of total 228 comparisons) where we failed to reject the null hypothesis across number pairs; intuitively, the models find it difficult to discriminate between these number pairs.\footnote{From this plot we excluded 24 comparisons for number pairs (1, 2) and (6, 7) since means within these pairs are the same by the design of networks' training objective.}
The figure shows that pairs at the higher end of the numerical range, such as (5, 7) and (5, 6) are frequently indistinguishable which is in contrast to the pairs on the lower end of the numerical scale.
We conclude that similar to the empirical data, \swin{} and \resnet{} better distinguish numerosities at the lower end of the number range compared to those of the higher end of number ranges. Moreover, this effect is stronger among \swin{} responses compared to \resnet{}, suggesting that number representations learned by \swin{} are more discernible.

\section{Discussion}

Number discrimination is a core aspect of basic numerical competence in humans and animals. We investigate if recent, state-of-the-art neural networks used in computer vision exhibit the capacity of discriminating between small and large quantities.
We evaluate these models on the numerical bisection task where models learn to categorize numerosity of sets of items, and we investigate their performance on novel stimuli and novel numerosities.

We find that \resnet{} and \swin{}, the two models with vision-specific inductive biases, achieve the smallest errors when categorizing novel stimuli as \emph{few} or \emph{many}.
Psychometric curves of models trained on a wide range of stimuli, as well as those of models trained and tested on the same type of stimuli, often resemble the response curves of animals and humans on the same task. In addition, \swin{} responses are more discernible for smaller numbers compared to larger numbers, and its internal representations are structured in a way that reflects number category and order. 
%
\swin's predecessor, \vit, which is also a transformer-based model, albeit with a weak image-specific inductive bias, has errors on the task that are comparable to or even higher than the basic, substantially smaller \mlp{} baseline. This is surprising considering that performance of \vit{} is within a few percentage points of \swin{} performance on different computer vision benchmarks~\cite{liu2021swin}.


Finally, when controlled for perceptual attributes (\eg{} keeping the total white area constant during training, but varying area during testing), most of these models show poor transfer of the number discrimination skill.
This might mean that models latch onto features that, while correlated with number, are considered non-numerical in the literature~\cite{Honig1989-dj,Testolin2020-tg}. When analyzing the internal representations of number in \swin{}, we find that often, despite poor transfer, number representations are structured in an interpretable way. In other words, although these representations could in theory support number discrimination, we do not observe this in practice.
%
One possible reason for poor transfer might be that the models are trained in a limited data regime, in contrast to humans and animals whose numerical cognition develops gradually in a rich environmental context, and who might be biologically predisposed to represent and process numerical quantities~\cite{dehaene1998abstract}. Future work should explore whether pretraining models on larger and more diverse sets of images would result in a more transferable skill.
%
%
Finally, we only investigate one specific task---the numerical bisection, and it remains to be explored whether our findings generalize across other perceptual domains.

\subsection{Acknowledgments}
The authors would like to thank Stephanie Chan for detailed feedback on this manuscript as well as other colleagues at DeepMind for feedback and discussions that helped improve this work.

\bibliographystyle{apacite}
\setlength{\bibleftmargin}{.125in}
\setlength{\bibindent}{-\bibleftmargin}
\bibliography{biblio}

\end{document}